\documentclass[letterpaper]{article} % DO NOT CHANGE THIS
\usepackage{aaai25}
\usepackage{times}  % DO NOT CHANGE THIS
\usepackage{helvet}  % DO NOT CHANGE THIS
\usepackage{courier}  % DO NOT CHANGE THIS
\usepackage[hyphens]{url}  % DO NOT CHANGE THIS
\usepackage{graphicx} % DO NOT CHANGE THIS
\usepackage{natbib}  % DO NOT CHANGE THIS AND DO NOT ADD ANY OPTIONS TO IT
\usepackage{caption} % DO NOT CHANGE THIS AND DO NOT ADD ANY OPTIONS TO IT
\usepackage{algorithm}
\usepackage{algorithmic}

\usepackage{booktabs}       % professional-quality tables
\usepackage{amsfonts}       % blackboard math symbols
\usepackage{amsmath}
\usepackage{xcolor}         % colors
\usepackage{multirow}
\usepackage{bm}

\newcommand{\update}[1]{{\color{black}{#1}}}
\usepackage{newfloat}
\usepackage{listings}
\DeclareCaptionStyle{ruled}{labelfont=normalfont,labelsep=colon,strut=off} % DO NOT CHANGE THIS
\floatstyle{ruled}
\newfloat{listing}{tb}{lst}{}
\floatname{listing}{Listing}
\nocopyright
\title{Wavelet-based Disentangled Adaptive Normalization for Non-stationary Times Series Forecasting}
\author{
  Junpeng Lin\textsuperscript{\rm 1},
  Tian Lan\textsuperscript{\rm 1},
  Bo Zhang\textsuperscript{\rm 1},
  Ke Lin\textsuperscript{\rm 2},
  Dandan Miao\textsuperscript{\rm 2},\\
  Huiru He\textsuperscript{\rm 3},
  Jiantao Ye\textsuperscript{\rm 3},
  Chen Zhang\textsuperscript{\rm 1}\thanks{Corresponding author},
  Yan-fu Li\textsuperscript{\rm 1}
}
\affiliations{
  \textsuperscript{\rm 1}Department of Industrial Engineering, Tsinghua University\\
  \textsuperscript{\rm 2}Shanghai Research Center, Huawei Technologies\\
  \textsuperscript{\rm 3}Songshan Lake Research Center, Huawei Technologies\\
  \{linjp22,lant23\}@mails.tsinghua.edu.cn, \{bz16328,zhangchen01,liyanfu\}@tsinghua.edu.cn, \{linke2,miaodandan,hehuiru,yejiantao\}@huawei.com
}
\usepackage{bibentry}
\begin{document}

\maketitle

\begin{abstract}
Forecasting non-stationary time series is a challenging task because their statistical properties often change over time, making it hard for deep models to generalize well. Instance-level normalization techniques can help address shifts in temporal distribution. However, most existing methods overlook the multi-component nature of time series, where different components exhibit distinct non-stationary behaviors. In this paper, we propose Wavelet-based Disentangled Adaptive Normalization (WDAN), a model-agnostic framework designed to address non-stationarity in time series forecasting. WDAN uses discrete wavelet transforms to break down the input into low-frequency trends and high-frequency fluctuations. It then applies tailored normalization strategies to each part. For trend components that exhibit strong non-stationarity, we apply first-order differencing to extract stable features used for predicting normalization parameters. Extensive experiments on multiple benchmarks demonstrate that WDAN consistently improves forecasting accuracy across various backbone model. Code is available at this repository: \url{https://github.com/MonBG/WDAN}.
\end{abstract}
% TODO


\section{Introduction}

Time series forecasting is an essential task in many fields, such as energy \cite{singh2013overview}, economics \cite{ahmadi2019presentation}, transportation \cite{yu2018spatio}, and healthcare \cite{kaushik2020ai}. While deep learning models have made significant progress in capturing temporal patterns \cite{zhou2022fedformer, zhang2022crossformer, Kitaev2020Reformer}, real-world time series remain difficult to predict due to their non-stationary characteristic. In non-stationary time series, statistical properties such as mean and variance change over time, which hinders generalization of deep learning models \cite{lu2018learning,li2022ddg}. While most current research focuses on capturing complex patterns through sophisticated architectures, the non-stationarity issue has not been fully addressed.

Our work is motivated by the fact that time series usually contain multiple components with different characteristics. For example, real-world sequences often include slow-moving trends, periodic patterns, and high-frequency noise \cite{rb1990stl}. Each of these components has its unique non-stationary behaviors and thus requires specialized handling. Traditional methods that treat the entire time series as a single entity often overlook these distinct behaviors. Time-frequency analysis provides an effective way to tackle this issue by breaking down a time series into separate components at different temporal scales, thereby revealing the underlying structure of non-stationarity.

\begin{figure}[h]
  \centering
  \includegraphics[width=1\linewidth]{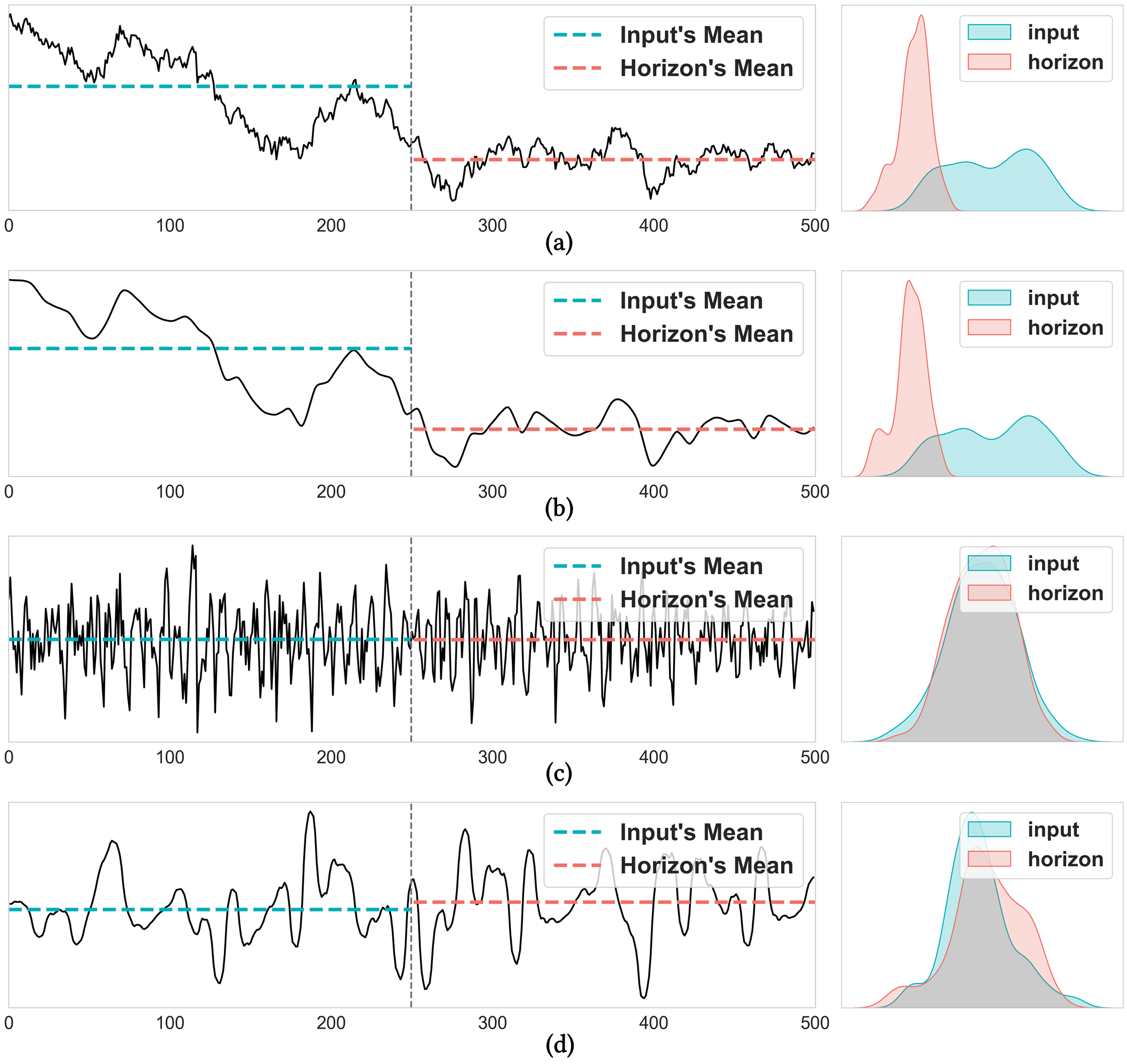}
  \caption{An illustration of a forecast sample indicating the changes of daily exchange rate \cite{lai2018modeling}. The vertical dotted line divides the input and horizon series. (a) The original series and corresponding distributions of input series and output series. (b) The low frequency trend component obtained by DWT and the corresponding distributions. (c) The residual component and the corresponding distributions. (d) The first-order trend difference and the corresponding distributions.}
  \label{fig:intro}
\end{figure}

Fig. \ref{fig:intro} presents a forecasting example to illustrate our opinions. In Fig. \ref{fig:intro}(a), the original time series clearly shows non-stationarity with changing mean and variance. We apply discrete wavelet transform \cite{chaovalit2011discrete,huang2002coiflet} to break down the series into a low-frequency trend component and a residual component containing several high-frequency parts. The non-stationarity is mainly captured by the slowly changing low-frequency trend series as shown in Fig. \ref{fig:intro}(b). We can see significant differences not just between the input and horizon distributions, but even between adjacent segments within the input series itself. In contrast, the residual component in Fig. \ref{fig:intro}(c) shows more stationary cyclical patterns. In \ref{fig:intro}(d), we additionally plot the first-order differenced series of the trend component, which appears to be more stationary compared to the original trend series. This finding suggests that the differenced features may facilitate the deep model in capturing the non-stationary dynamics of the trend sequence. Conventional normalization methods typically apply the same transformation to the entire sequence, thereby failing to capture the distinct behaviors of different components. Such limitations motivates the need for a more fine-grained normalization approach that leverages the inherent structure of time series.

% TODO: the figure needs to be corrected

Researchers have tried various approaches to handle non-stationarity, including differencing, seasonal decomposition and domain-specific feature engineering \cite{box2015time,liu2023koopa}. Recently, normalization techniques \cite{kim2021reversible,fan2023dish,liu2023adaptive,dai2024ddn} have gained popularity due to their simplicity and compatibility with many deep learning models. Instance normalization methods help reduce distribution shifts by adjusting input statistics during training and testing. However, current normalization methods extract statistics directly from the raw sequence without considering the multi-component nature of time series. As a result, the statistics obtained blend information from components with different characteristics, which can lead to information loss during normalization process.

To overcome these limitations, we propose Wavelet-based Disentangled Adaptive Normalization (WDAN), a model-agnostic framework for non-stationary time series forecasting. WDAN starts by applying discrete wavelet transforms to decompose each input sequence into trend and residual components. We then extract normalization statistics separately for each component to reflect their unique statistical properties. For the trend component, which usually shows stronger non-stationarity, we apply first-order differencing to obtain more stable features for statistics prediction. A lightweight prediction module estimates future normalization statistics, allowing us to adaptively de-normalize the model outputs.

Our main contributions are summarized as follows: (\textbf{i}) We introduce a novel perspective on non-stationarity by identifying and separately modeling the distinct statistical behaviors of different time series components through wavelet decomposition. (\textbf{ii}) We propose WDAN, a model-agnostic normalization framework that disentangles non-stationary components and applies specialized treatments to each, including first-order differencing for trend modeling. (\textbf{iii}) Extensive experiments on benchmark datasets demonstrate that WDAN consistently improves forecasting accuracy across multiple backbone models, with particularly significant gains in highly non-stationary settings.

\section{Related Work}

\subsection{Time Series Forecasting}

As an important issue in many fields such as finance, energy and transportation, time series forecasting has been studied for decades. Classical statistical models including ARIMA \cite{box2015time} use linear autoregressive structures to capture temporal dependencies. However, despite their simplicity and interpretability, statistical models are limited by strong assumptions about the ideal properties of the data.

Recently developed deep learning-based models have achieved impressive performance by leveraging non-linear modeling capabilities. Recurrent neural networks (RNNs) \cite{hochreiter1997long, cho2014learning, chung2015recurrent}, temporal convolutional networks (TCNs) \cite{bai2018empirical, sen2019think}, and Transformer-based architectures \cite{vaswani2017attention, zhou2021informer, zhou2022fedformer} have been widely adopted for univariate and multivariate time series forecasting. These models vary in how they capture temporal dependencies and long-term trends. Some works further improve forecasting performance by incorporating time series decomposition \cite{wu2021autoformer, zeng2023transformers} or time-frequency representations \cite{zhou2022fedformer}.

While many recent models focus on modeling cross-variable dependencies in multivariate settings using attention mechanisms \cite{zhang2022crossformer, liu2023itransformer} or graph neural networks \cite{bai2020adaptive, wu2020connecting}, our work takes a channel-independent approach. This choice is supported by previous studies showing that channel-independent models are more robust to distribution shifts and non-stationary conditions~\cite{wen2023onenet, han2024capacity}.

\subsection{Non-stationary Time Series Forecasting}

Non-stationarity is a key challenge in time series forecasting, as distributional shifts over time can lead to poor generalization and unstable predictions. Pre-process the input series with stationarization methods is a straightforward approach to tackle the non-stationarity. Traditional statistical models such as ARIMA \cite{box2015time} apply differencing to remove trends and achieve stationarity. In deep learning, normalization is the most popular technique to deal with non-stationarity. RevIN \cite{kim2021reversible} proposes a symmetric normalization scheme that normalizes input series and then denormalizes model outputs using instance normalization. DAIN \cite{passalis2019deep} employs a nonlinear transformation on normalization statistics to adaptively stationarize inputs. However, the above methods overlook the distribution discrepancy between the input series and the horizon series. Dish-TS \cite{fan2023dish} proposes to address this issue by learning a mapping from the input distribution to the output distribution characterized by means and variances. SAN \cite{liu2023adaptive} notices the distribution shifts across compact time slices and proposes to learn the distribution mapping on time slice level. While these methods improve robustness to distribution shifts, they typically apply the normalization scheme to the entire sequence, overlooking the fact that real-world time series often contain multiple components with different behaviors.

Other approaches model non-stationarity from a dynamical systems perspective. For example, Koopman-based methods \cite{liu2023koopa, wang2022koopman} treat non-stationary time series as a non-linear dynamical system and transform the system into a measurement function space, which can be described by a linear Koopman operator. While theoretically sound, these approaches often require specific domain knowledge to define appropriate functions or operators.

Our approach differs from existing methods by explicitly recognizing that different components of time series show different types and degrees of non-stationarity. By using wavelet decomposition to disentangle these components and applying tailored normalization to each, our approach addresses a fundamental limitation in current normalization techniques. Moreover, our framework incorporates differencing methods for handling strongly non-stationary trend components, while maintaining the model-agnostic advantages of normalization-based approaches.

\section{Methodology}

We proposes the \textbf{W}avelet-based \textbf{D}isentangled \textbf{A}daptive \textbf{N}ormalization (WDAN) framework for non-stationary multivariate time series forecasting. Given the input series $\boldsymbol{X}\in\mathbb{R}^{N\times T}=(\boldsymbol{x}^1, ..., \boldsymbol{x}^N)$ with $N$ variables and $T$ input time steps, we aim to predict the future series $\boldsymbol{Y}\in\mathbb{R}^{N\times H}(\boldsymbol{y}^1, ..., \boldsymbol{y}^N)$ with prediction horizon length $H$. The core idea lies in disentangling the non-stationary components via wavelet transform and adaptively predict future non-stationary normalization statistics from the decomposed series. In this section, we will present the detailed pipeline of the entire framework and elaborate how to address the non-stationarity problem during time series forecasting.

\subsection{Overall Framework}

\begin{figure*}[htbp]
    \centering
    \includegraphics[width=1\linewidth]{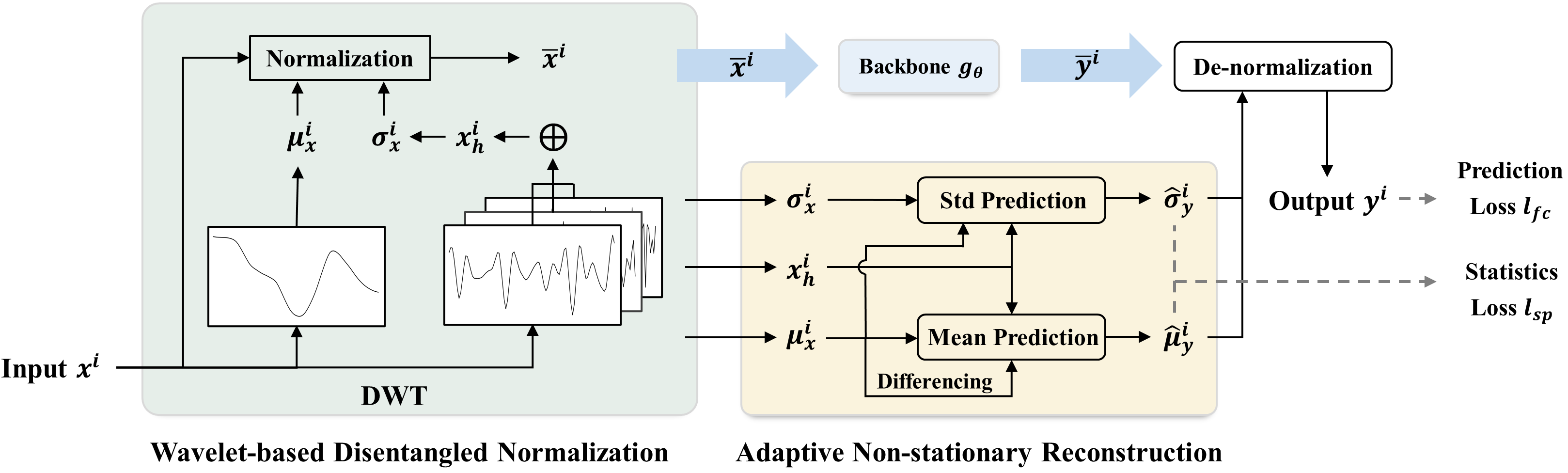}
    \caption{The illustration of the proposed WDAN framework}
    \label{fig:wdan_framework}
\end{figure*}

As a model-agnostic framework, WDAN adopts a popular normalization framework, which first stationarizes the input sequences by normalization and later restores the non-stationarity information by denormalizing the output sequences. Unlike the existing normalization methods, WDAN does not directly compute normalization statistics on the original input sequences. Instead, as shown in Fig. \ref{fig:wdan_framework}, WDAN begins by disentangling the input sequences into low-frequency and high-frequency components via discrete wavelet transform (DWT), which captures persistent trends and transient fluctuations, respectively. These components then undergo instance-specific normalization, where low-frequency signals guide the estimation of evolving mean trends, and high-frequency residuals inform time-varying variance calibration. The normalized sequence is input to a backbone forecasting model, i.e., RNNs or Transformers, to obtain the normalized prediction sequences. An statistics prediction module subsequently predicts future distribution statistics for denormalization by synthesizing multi-scale dynamics, including the differential trends. To ensure stable model training, the framework adopts a two-stage training strategy.

\subsection{Wavelet-based Disentangled Normalization}

We first decompose each input series $\boldsymbol{x}^i$ into low-frequency appriximation coefficients $\boldsymbol{c}_l^i$ and high-frequency detail coefficients $\{\boldsymbol{c}_{h_k}^i\}_{k=1}^K$ by DWT of level $K$. The coefficients are reconstructed into interpretable temporal components via inverse DWT. The procedure is formulated as follows:
\begin{align}
  &\boldsymbol{c}_{l_k}^i,\boldsymbol{c}_{h_k}^i=\text{DWT}_{\phi_{l,h}}(\boldsymbol{c}_{l_{k-1}}^i), \forall k=1,...,K\\
  &\boldsymbol{c}_{l_0}^i=\boldsymbol{x}^i,\quad\boldsymbol{c}_{l}^i=\boldsymbol{c}_{l_K}^i,\\
  &\boldsymbol{x}_l^i=\text{IDWT}(\boldsymbol{c}_l^i,\boldsymbol{0}),\quad \boldsymbol{x}_h^i=\sum_{k=1}^K\text{IDWT}(\boldsymbol{0}, \boldsymbol{c}_{h_k}^i),
\end{align}
where $\phi_{l,h}$ is a pair of wavelet bases. \update{$\boldsymbol{x}_l^i$ and $\boldsymbol{x}_h^i$ represent the low-frequency trend component and the high-frequency residual component of $\boldsymbol{x}^i$, respectively.} Unlike Fourier-based methods that impose fixed frequency assumptions, wavelet decomposition adaptively isolates non-stationary patterns while preserving localized temporal features. $\boldsymbol{x}_l^i$ captures the continuously changing trends, which is the main source of non-stationarity. $\boldsymbol{x}_h^i$ aggregates transient fluctuations across different frequency bands, which reflects the inherent patterns of the input series.

Normalization parameters are dynamically derived from these components. \update{Since the low-frequency signal provides the time-varying means due to non-stationarity, we define the mean series for input sequence normalization as $\boldsymbol{\mu}_x^i=(\mu_x^i[t-T+1],\dots,\mu_x^{i}[t])\in\mathbb{R}^T$, where $\mu_x^i[t]=x^i_l[t]$ and $x_l^i[t]$ denotes the value of $\boldsymbol{x}_l^i$ at time step $t$}. 
% \chen{this seems not correct} 
High-frequency residuals encompass both the localized features of the inherent patterns of the input series and the non-stationary variations caused by abrupt changes, noise, etc. Therefore, we derive the point-level standard deviation parameters from $\boldsymbol{x}_h^i$ by a sliding window along the temporal dimension. Replication padding is added to align the length of sliding statistics to the original series. This process is formulated as follows:\update{
\begin{align}
  \mu^i_h[t]&=\frac{1}{2w+1}\sum_{j=-w}^{w}x_h^i[t+j],\\
  \left(\sigma_x^i[t]\right)^2&=\frac{1}{2w+1}\sum_{j=-w}^{w}\left(x_h^i[t+j]-\mu_h^i[t]\right)^2,
\end{align}
% \chen{Eq. 4 seems incorrect, what is $\mu_h$?}
where $2w+1$ is the size of the sliding window. Thus, the point-level instance normalization process is as follows:
\begin{equation}
  \bar{x}^i[t]=\frac{1}{\sigma_x^i[t]+\epsilon}(x^i[t]-\mu_x^i[t]).
\end{equation}}
Here, $\bar{\boldsymbol{x}}^i$ is the stationary series and $\epsilon$ is a small positive number to prevent the denominator from zero. Through series decomposition, we provide a more detailed explanation of the sources of non-stationarirty and then extract the normalization parameters from different components. The normalized series $\bar{\boldsymbol{X}}=(\bar{\boldsymbol{x}}^1,...,\bar{\boldsymbol{x}}^N)$ is input to the downstream backbone forecasting model $g_\theta$.

\subsection{Adaptive Non-stationarity Reconstruction}

To compensate for evolving distribution shifts, we predict future distribution parameters by fusing multi-scale dynamics. Then we denormalize the output of backbone forecasting models with the predicted statistics. 

\subsubsection{Statistics Prediction with Differencing}

Following previous distribution prediction works, we use a multi-layer perceptron network to predict future distributions efficiently. Instead of learning the statistics directly, we adopt a residual learning technique to predict the difference between future statistics and the overall mean of the current statistics series. We define
\begin{equation}
  \bar{\mu}^i=\frac{1}{T}\sum_{t=1}^{T}\mu_x^i[t],\quad \bar{\sigma}^i=\frac{1}{T}\sum_{t=1}^{T}\sigma_x^i[t].
\end{equation}
Subsequently, the difference between the statistics series and the overall mean is input into the prediction module. Considering the potential interaction between high- and low-frequency signals, we take both high and low frequency signals as input. Furthermore, due to the non-stationarity of the mean sequence, we additionally perform first-order differencing on it and input the differenced series into the mean prediction module. The overall procedure is formulated as follows:
\update{
\begin{align}
  &\quad\quad\quad\quad\quad\quad\quad\tilde{\mu}_x^i[t]=\mu_x^i[t]-\bar{\mu}_x^i[t],\\
  &\quad\quad\quad\quad\quad\quad\quad\tilde{\sigma}_x^i[t]=\sigma_x^i[t]-\bar{\sigma}_x^i[t],\\
  &\quad\quad\quad\quad\quad\quad\ \Delta\tilde{\mu}_x^i[t]=\tilde{\mu}_x^i[t]-\tilde{\mu}_x^i[t-1],\\
  &\hat{\boldsymbol{\mu}}^i_y=\text{MLP}_\mu\left(
    \text{MLP}_1(\tilde{\boldsymbol{\mu}}_x^i)\ \|\ \text{MLP}_2(\Delta\tilde{\boldsymbol{\mu}}_x^i)\ \|\ \text{MLP}_3(\boldsymbol{x}_h^i)
    \right)+\bar{\mu}^i,\\
  &\hat{\boldsymbol{\sigma}}^i_y=\text{MLP}_\sigma\left(
    \text{MLP}_4(\tilde{\boldsymbol{\sigma}}_x^i)\ \|\ \text{MLP}_1(\tilde{\boldsymbol{\mu}}_x^i)\ \|\ \text{MLP}_3(\boldsymbol{x}_h^i)
    \right)+\bar{\sigma}^i,
\end{align}
}
% \chen{should it be $\mu^{i}[t]-\bar{\mu}^{i}$, also for $sigma$, bold, (7) use [t], define muy}
where $\|$ denotes concatenation. \update{The statistics prediction module makes predictions at the point level, meaning that the parameter series for output sequence de-normalization $\hat{\boldsymbol{\mu}}^i_y\in\mathbb{R}^{H}$ and $\hat{\sigma}^i_y\in\mathbb{R}^H$. $\text{MLP}_i:\mathbb{R}^T\to\mathbb{R}^D$ for $i=1,2,3,4$ initially maps the input sequence along the temporal dimension into a high-dimensional latent vector, where $D$ is the hidden dimension. Then $\text{MLP}_\mu$ and $\text{MLP}_\sigma$ project the concatenated feature vectors into predictions of the parameter series for de-normalization.}

\subsubsection{De-normalization}

After the aforementioned statistics predictions, WDAN denormalizes the output sequence of the backbone forecasting model to restore the non-stationary factors. Specifically, WDAN performs instance-specific denormalization as follows:
\begin{align}
  &\bar{\boldsymbol{Y}}=g_\theta(\bar{\boldsymbol{X}}),\quad \bar{\boldsymbol{Y}}=(\bar{\boldsymbol{y}}^1,...,\bar{\boldsymbol{y}}^N),\\
  &\hat{\boldsymbol{y}}^i=\bar{\boldsymbol{y}}^i\odot(\hat{\boldsymbol{\sigma}}^i_y+\epsilon)+\hat{\boldsymbol{\mu}}^i_y,
\end{align}
where $\odot$ is the elementwise product. Finally, we obtain the final prediction $\hat{\boldsymbol{Y}}=(\hat{\boldsymbol{y}}^1,...,\hat{\boldsymbol{y}}^N)$ of the whole framework.

\subsection{Three-stage Training Strategy}
\label{sec:training}

The training process constitutes a bi-level optimization problem where the results of statistical prediction greatly impact the prediction performance of the overall framework. We adopt a three-stage training strategy to decouple statistics prediction from temporal pattern learning of the backbone model. In the first training stage, the statistics prediction network undergoes dedicated pretraining using ground truth statistics series, i.e.,
\begin{equation}
\theta_d=\arg\min_{\theta_d}\sum_il_{sp}\left((\hat{\boldsymbol{\mu}}_y^i,\hat{\boldsymbol{\sigma}}^i_y),(\boldsymbol{\mu}_y^i,\boldsymbol{\sigma}_y^i),\theta_d\right),
\end{equation}
where $\theta_d$ represents the parameters of the statistics prediction module. $l_{sp}$ is a loss function between predicted statstics and ground truth, e.g., mean absolute error (MAE) or mean squared error (MSE). The second training stage freezes the pretrained statistics prediction module and focuses on optimizing the backbone forecasting model $g_\theta$, i.e.,
\begin{equation}
    \theta_g=\arg\min_{\theta_g}\sum_il_{fc}\left(
      \hat{\boldsymbol{y}}^i,\boldsymbol{y}^i,\theta_g
    \right),
\end{equation}
where $\theta_g$ represents the parameters of $g_\theta$. $l_{fc}$ if a loss function to evaluate the overall forecasting results. The third training stage, considering the mutual influence between the backbone model and the statistics prediction module, we perform joint fine-tuning with a relatively smaller learning rate, i.e.,
\begin{equation}
    \{\theta_d, \theta_g\}=\arg\min_{\theta_g}\sum_il_{fc}\left(
      \hat{\boldsymbol{y}}^i,\boldsymbol{y}^i,\{\theta_d, \theta_g\}
    \right).
\end{equation}

\section{Experiments}

In this section, we evaluate the effectiveness of the proposed WDAN framework on multiple benchmark datasets for multivariate time series forecasting.

\subsection{Experimental Setup}

\subsubsection{Datasets}
\label{sec:datasets}

To demonstrate the effectiveness of WDAN in handling non-stationarity across various data scenarios, we conduct experiments on three widely-used benchmark datasets: (1) \textbf{Exchange} \cite{lai2018modeling} dataset, which collects the daily exchange rate of 8 countries from 1990 to 2016. (2) \textbf{ETT} \cite{zhou2021informer} dataset, which records oil temperature and load features of electricity transformers from July 2016 to July 2018. It consists of four sub-datasets, where ETTh is sampled hourly and ETTm is sampled every 15 minutes. (3) \textbf{Weather}\footnote{\url{https://www.bgc-jena.mpg.de/wetter/}} dataset, which consists of 21 weather indicators including air temperature and humidity collected every 10 minutes in 2021. (4) \textbf{Electricity}\footnote{\url{https://archive.ics.uci.edu/ml/datasets/ElectricityLoadDiagrams20112014}}, which contains the electricity consumption data of 321 clients from July 2016 to July 2019. Detailed information about these datasets is listed in Table \ref{tab:dataset}, where we also report the results of the Augmented Dickey-Fuller (ADF) test that assesses time series stationarity. A smaller absolute value of the ADF statistic indicates stronger non-stationarity.

Following previous works, we split each dataset into training, validation, and test sets in chronological order. The split ratio is 6:2:2 for ETT datasets and 7:1:2 for other datasets. Additionally, we perform z-score normalization on the datasets based on the training data statistics as preprocessing, which allows measurements of different variables on the same scale. Note that z-score normalization uses global statistics for normalization, meaning it cannot handle non-stationarity since the statistics used in the normalization process are fixed across different samples.

\begin{table}[htbp]
    \centering
    \begin{tabular}{ccccc}
    \toprule
    Dataset & Vars & \begin{tabular}[c]{@{}c@{}}Sampling\\ Frequency\end{tabular} & \begin{tabular}[c]{@{}c@{}}Sequence\\ Length\end{tabular} & \begin{tabular}[c]{@{}c@{}}ADF\\ Statistics\end{tabular} \\
    \midrule
    Exchange & 8 & 1 day & 7588 & -1.90 \\
    ETTh1 & 7 & 1 hour & 17420 & -5.91 \\
    ETTh2 & 7 & 1 hour & 17420 & -4.13 \\
    ETTm1 & 7 & 15 mins & 69680 & -14.98 \\
    ETTm2 & 7 & 15 mins & 69680 & -5.66 \\
    Weather & 21 & 10 mins & 52696 & -26.68 \\
    Electricity & 321 & 1 hour & 26304 & -8.44 \\
    \bottomrule
    \end{tabular}
    \caption{Statistics of benchmark datasets}
    \label{tab:dataset}
\end{table}

\subsubsection{Baseline Methods}

WDAN is a model-agnostic approach that can be applied to any mainstream time series forecasting model. To demonstrate its versatility, we integrate WDAN with multiple representative models, including previously proposed FEDformer \cite{zhou2022fedformer}, as well as recently introduced Crossformer \cite{zhang2022crossformer}, PatchTST \cite{Yuqietal-2023-PatchTST}, and iTransformer \cite{liu2023itransformer}.

To highlight WDAN's effectiveness in handling non-stationarity, we also compare it with mainstream normalization frameworks, including: (1) SAN \cite{liu2023adaptive}: adopts segment-level instance normalization and predicts future statistics through segmented sequence prediction; (2) DDN \cite{dai2024ddn}: employs point-level instance normalization and normalizes the entire sequence in both time and frequency domains.

\subsubsection{Implementation Details}
\label{sec:main_imp}

We use Mean Squared Error (MSE) as the loss function across all experiments. For evaluation metrics, we report both MSE and Mean Absolute Error (MAE) on the test set, where lower MSE/MAE indicates better predictive performance. Consistent with existing works, all models are trained and tested on four different prediction horizons $H \in \{96, 192, 336, 720\}$, with a fixed historical sequence length of $L=720$. We report the average results of three runs for each model on each dataset to ensure stable performance. More implementation details of our experiments can be referred to the appendix.

\subsection{Main Results}

Table \ref{tab:main_result} presents the multivariate time series forecasting performance of WDAN applied to four mainstream time series forecasting models. The results clearly show that WDAN significantly improves the predictive performance of all backbone models in most cases, demonstrating the effectiveness and universality of the proposed method in handling non-stationary time series data. WDAN delivers consistent performance improvements across all backbone models at different prediction horizons.

Notably, as the prediction length increases from $H=96$ to $H=720$, the magnitude of performance improvement provided by WDAN becomes more pronounced, indicating that non-stationarity handling methods have greater advantages for long-term forecasting. This may be because distributional shifts in sequences typically occur gradually, making non-stationarity more significant for long-term predictions. For example, when $H=720$, WDAN reduces iTransformer's MSE and MAE on the ETTh1 dataset by 19.6\% and 13.8\% respectively, while at $H=96$, the reduction is only 5.6\% and 5.9\%.

Furthermore, since iTransformer and PatchTST already incorporate RevIN, a basic instance normalization method for handling non-stationarity in their original models, the performance improvement provided by WDAN on these two models is relatively smaller compared to its impact on Crossformer and FEDformer. Nevertheless, when we replace RevIN with WDAN in iTransformer and PatchTST, our method still achieves better performance, indicating that WDAN is superior to RevIN in handling non-stationarity. Additionally, the application of WDAN to Crossformer and FEDformer compensates for the limitations of these relatively earlier models in non-stationary modeling, enabling them to perform comparably to state-of-the-art models like iTransformer, further highlighting the importance of non-stationarity handling in sequence forecasting.

\begin{table*}[!t]
  \centering
%   \scalebox{0.8}{
    {\setlength{\tabcolsep}{1mm}
  \begin{tabular}{c|c|cccc|cccc|cccc|cccc}
  \toprule
  \multicolumn{2}{c|}{Methods} & \multicolumn{2}{c}{iTransformer} & \multicolumn{2}{c|}{+WDAN} & \multicolumn{2}{c}{PatchTST} & \multicolumn{2}{c|}{+WDAN} & \multicolumn{2}{c}{Crossformer} & \multicolumn{2}{c|}{+WDAN} & \multicolumn{2}{c}{FEDformer} & \multicolumn{2}{c}{+WDAN} \\
  \cmidrule(r){3-4}  \cmidrule(r){5-6} \cmidrule(r){7-8} \cmidrule(r){9-10} \cmidrule(r){11-12} \cmidrule(r){13-14} \cmidrule(r){15-16} \cmidrule(r){17-18}
  \multicolumn{2}{c|}{Metric} & MSE & MAE & MSE & MAE & MSE & MAE & MSE & MAE & MSE & MAE & MSE & MAE & MSE & MAE & MSE & MAE \\
  \midrule
  \multirow{4}{*}{\rotatebox[origin=c]{90}{Exchange}} & 96 & 0.118 & 0.250 & \textbf{0.083} & \textbf{0.203} & 0.101 & 0.230 & \textbf{0.084} & \textbf{0.204} & 0.561 & 0.547 & \textbf{0.084} & \textbf{0.205} & 0.744 & 0.668 & \textbf{0.087} & \textbf{0.205} \\
 & 192 & 0.246 & 0.370 & \textbf{0.176} & \textbf{0.300} & 0.200 & 0.326 & \textbf{0.176} & \textbf{0.302} & 0.798 & 0.683 & \textbf{0.174} & \textbf{0.300} & 0.928 & 0.760 & \textbf{0.180} & \textbf{0.305} \\
 & 336 & 0.429 & 0.493 & \textbf{0.361} & \textbf{0.436} & 0.405 & 0.470 & \textbf{0.347} & \textbf{0.429} & 1.455 & 0.975 & \textbf{0.350} & \textbf{0.431} & 1.180 & 0.857 & \textbf{0.386} & \textbf{0.456} \\
 & 720 & 1.101 & 0.796 & \textbf{1.009} & \textbf{0.767} & 1.463 & 0.904 & \textbf{1.081} & \textbf{0.781} & 2.050 & 1.163 & \textbf{1.019} & \textbf{0.762} & 1.901 & 1.068 & \textbf{1.110} & \textbf{0.789} \\
 \midrule
\multirow{4}{*}{\rotatebox[origin=c]{90}{ETTh1}} & 96 & 0.390 & 0.422 & \textbf{0.368} & \textbf{0.397} & 0.399 & 0.419 & \textbf{0.369} & \textbf{0.398} & 0.391 & 0.431 & \textbf{0.366} & \textbf{0.396} & 0.486 & 0.498 & \textbf{0.381} & \textbf{0.415} \\
 & 192 & 0.427 & 0.448 & \textbf{0.406} & \textbf{0.421} & 0.447 & 0.450 & \textbf{0.407} & \textbf{0.422} & 0.490 & 0.493 & \textbf{0.401} & \textbf{0.415} & 0.522 & 0.512 & \textbf{0.417} & \textbf{0.438} \\
 & 336 & 0.459 & 0.471 & \textbf{0.428} & \textbf{0.444} & 0.472 & 0.474 & \textbf{0.443} & \textbf{0.449} & 0.939 & 0.758 & \textbf{0.420} & \textbf{0.429} & 0.545 & 0.534 & \textbf{0.436} & \textbf{0.452} \\
 & 720 & 0.562 & 0.545 & \textbf{0.452} & \textbf{0.470} & 0.519 & 0.511 & \textbf{0.515} & \textbf{0.504} & 1.080 & 0.796 & \textbf{0.446} & \textbf{0.468} & 0.678 & 0.620 & \textbf{0.473} & \textbf{0.484} \\
 \midrule
\multirow{4}{*}{\rotatebox[origin=c]{90}{ETTh2}} & 96 & 0.301 & 0.360 & \textbf{0.269} & \textbf{0.334} & 0.322 & 0.376 & \textbf{0.271} & \textbf{0.335} & 1.225 & 0.764 & \textbf{0.269} & \textbf{0.336} & 0.405 & 0.459 & \textbf{0.277} & \textbf{0.342} \\
 & 192 & 0.379 & 0.408 & \textbf{0.331} & \textbf{0.376} & 0.412 & 0.432 & \textbf{0.336} & \textbf{0.376} & 1.253 & 0.810 & \textbf{0.333} & \textbf{0.378} & 0.431 & 0.480 & \textbf{0.347} & \textbf{0.386} \\
 & 336 & 0.415 & 0.436 & \textbf{0.358} & \textbf{0.403} & 0.450 & 0.455 & \textbf{0.357} & \textbf{0.401} & 1.500 & 0.920 & \textbf{0.359} & \textbf{0.405} & 0.436 & 0.481 & \textbf{0.391} & \textbf{0.423} \\
 & 720 & 0.435 & 0.463 & \textbf{0.377} & \textbf{0.429} & 0.497 & 0.486 & \textbf{0.394} & \textbf{0.440} & 3.825 & 1.606 & \textbf{0.384} & \textbf{0.428} & 0.502 & 0.514 & \textbf{0.483} & \textbf{0.478} \\
 \midrule
\multirow{4}{*}{\rotatebox[origin=c]{90}{ETTm1}} & 96 & 0.315 & 0.368 & \textbf{0.291} & \textbf{0.350} & 0.296 & 0.354 & \textbf{0.293} & \textbf{0.348} & 0.348 & 0.389 & \textbf{0.293} & \textbf{0.349} & 0.435 & 0.457 & \textbf{0.295} & \textbf{0.354} \\
 & 192 & 0.345 & 0.386 & \textbf{0.330} & \textbf{0.375} & 0.346 & 0.388 & \textbf{0.335} & \textbf{0.376} & 0.360 & 0.395 & \textbf{0.332} & \textbf{0.373} & 0.432 & 0.455 & \textbf{0.330} & \textbf{0.378} \\
 & 336 & 0.379 & 0.407 & \textbf{0.362} & \textbf{0.396} & 0.400 & 0.421 & \textbf{0.362} & \textbf{0.397} & 0.586 & 0.573 & \textbf{0.360} & \textbf{0.394} & 0.454 & 0.470 & \textbf{0.370} & \textbf{0.403} \\
 & 720 & 0.443 & 0.444 & \textbf{0.412} & \textbf{0.422} & 0.477 & 0.464 & \textbf{0.413} & \textbf{0.411} & 0.977 & 0.787 & \textbf{0.413} & \textbf{0.411} & 0.499 & 0.489 & \textbf{0.420} & \textbf{0.425} \\
 \midrule
\multirow{4}{*}{\rotatebox[origin=c]{90}{ETTm2}} & 96 & 0.182 & 0.276 & \textbf{0.162} & \textbf{0.253} & 0.179 & 0.272 & \textbf{0.172} & \textbf{0.257} & 0.310 & 0.388 & \textbf{0.162} & \textbf{0.254} & 0.317 & 0.379 & \textbf{0.167} & \textbf{0.258} \\
 & 192 & 0.244 & 0.317 & \textbf{0.217} & \textbf{0.292} & 0.248 & 0.317 & \textbf{0.240} & \textbf{0.304} & 0.387 & 0.438 & \textbf{0.216} & \textbf{0.290} & 0.338 & 0.390 & \textbf{0.237} & \textbf{0.299} \\
 & 336 & 0.298 & 0.351 & \textbf{0.273} & \textbf{0.331} & 0.311 & 0.357 & \textbf{0.282} & \textbf{0.334} & 0.614 & 0.558 & \textbf{0.278} & \textbf{0.336} & 0.366 & 0.404 & \textbf{0.282} & \textbf{0.338} \\
 & 720 & 0.376 & 0.401 & \textbf{0.353} & \textbf{0.382} & \textbf{0.418} & \textbf{0.426} & 0.442 & 0.432 & 1.279 & 0.813 & \textbf{0.390} & \textbf{0.405} & 0.421 & 0.446 & \textbf{0.395} & \textbf{0.417} \\
 \midrule
\multirow{4}{*}{\rotatebox[origin=c]{90}{Weather}} & 96 & 0.177 & 0.229 & \textbf{0.147} & \textbf{0.199} & 0.153 & 0.208 & \textbf{0.148} & \textbf{0.202} & 0.173 & 0.244 & \textbf{0.147} & \textbf{0.198} & 0.317 & 0.373 & \textbf{0.153} & \textbf{0.207} \\
 & 192 & 0.226 & 0.268 & \textbf{0.196} & \textbf{0.248} & 0.206 & 0.261 & \textbf{0.194} & \textbf{0.248} & 0.218 & 0.288 & \textbf{0.195} & \textbf{0.248} & 0.347 & 0.397 & \textbf{0.193} & \textbf{0.248} \\
 & 336 & 0.285 & 0.309 & \textbf{0.242} & \textbf{0.282} & 0.243 & \textbf{0.286} & \textbf{0.242} & 0.290 & 0.266 & 0.328 & \textbf{0.244} & \textbf{0.287} & 0.354 & 0.392 & \textbf{0.247} & \textbf{0.293} \\
 & 720 & 0.359 & 0.361 & \textbf{0.312} & \textbf{0.337} & 0.315 & 0.336 & \textbf{0.306} & \textbf{0.332} & 0.334 & 0.380 & \textbf{0.310} & \textbf{0.336} & 0.400 & 0.421 & \textbf{0.313} & \textbf{0.341} \\
 \midrule
\multirow{4}{*}{\rotatebox[origin=c]{90}{Electricity}} & 96 & 0.135 & 0.232 & \textbf{0.128} & \textbf{0.225} & 0.137 & 0.241 & \textbf{0.128} & \textbf{0.224} & 0.137 & 0.238 & \textbf{0.128} & \textbf{0.223} & 0.227 & 0.339 & \textbf{0.142} & \textbf{0.245} \\
 & 192 & 0.157 & 0.253 & \textbf{0.146} & \textbf{0.242} & 0.153 & 0.256 & \textbf{0.145} & \textbf{0.240} & 0.157 & 0.258 & \textbf{0.148} & \textbf{0.244} & 0.231 & 0.343 & \textbf{0.159} & \textbf{0.264} \\
 & 336 & 0.171 & 0.269 & \textbf{0.156} & \textbf{0.256} & 0.169 & 0.272 & \textbf{0.160} & \textbf{0.258} & 0.193 & 0.294 & \textbf{0.164} & \textbf{0.262} & 0.259 & 0.366 & \textbf{0.174} & \textbf{0.279} \\
 & 720 & 0.196 & 0.290 & \textbf{0.182} & \textbf{0.281} & 0.205 & 0.302 & \textbf{0.194} & \textbf{0.291} & 0.281 & 0.358 & \textbf{0.196} & \textbf{0.294} & 0.297 & 0.393 & \textbf{0.203} & \textbf{0.309} \\
   \bottomrule
  \end{tabular}
  }
  \caption{Multivariate time series forecasting performance of WDAN across different backbone models. The best results for each backbone model are highlighted in bold.}
  \label{tab:main_result}
\end{table*}

\subsection{Comparison with Normalization Methods}

To further validate WDAN's advantages in handling non-stationary time series data, Table \ref{tab:normalization_simple} shows a comparison of WDAN with mainstream normalization methods across different backbone models for multivariate time series forecasting. Table \ref{tab:normalization_simple} provides an overview of the average predictive performance across different prediction horizons, while more detailed results by prediction lengths are presented in the appendix.

From an overall performance perspective, WDAN achieves the best results on most dataset and model combinations. From the dataset dimension, WDAN demonstrates the most significant performance improvement on the Exchange and ETTh2 dataset, which exhibits the most pronounced non-stationarity (determined by the ADF test), achieving the best results in almost all prediction scenarios across all backbone models. On datasets with relatively weaker non-stationarity, such as ETTm1 and Weather, WDAN still shows stronger comprehensive predictive performance and competitive results in specific prediction scenarios compared to mainstream normalization methods. This not only demonstrates the versatility of non-stationarity handling methods across different data scenarios but also validates WDAN's advantages in handling non-stationarity.

Overall, the comparative experiments on normalization methods prove that WDAN's strategy of disentangling non-stationary components through wavelet decomposition and capturing dynamic changes through differencing enables more accurate handling of complex non-stationary patterns in time series. Compared to existing methods, WDAN significantly enhances the predictive capability for non-stationary time series data while maintaining the flexibility of the normalization framework.

\begin{table*}[!t]
\centering
\begin{tabular}{cc|ccccccc}
\toprule
\multicolumn{2}{c|}{Methods} & Exchange & ETTh1 & ETTh2 & ETTm1 & ETTm2 & Weather & Electricity \\
\midrule
\multirow{4}{*}{iTransformer} & +WDAN & \textbf{0.407} & \textbf{0.413} & \textbf{0.334} & \textbf{0.349} & 0.251 & \textbf{0.224} & 0.153 \\
 & +SAN & 0.416 & 0.433 & 0.358 & 0.350 & 0.268 & 0.226 & 0.155 \\
 & +DDN & 0.429 & 0.457 & 0.347 & 0.360 & \textbf{0.251} & 0.231 & \textbf{0.153} \\
 & IMP(\%) & 2.16 & 4.63 & 3.73 & 0.35 & -0.12 & 0.95 & -0.02 \\
 \midrule
\multirow{4}{*}{PatchTST} & +WDAN & \textbf{0.422} & \textbf{0.434} & \textbf{0.340} & \textbf{0.351} & \textbf{0.284} & \textbf{0.222} & \textbf{0.157} \\
 & +SAN & 0.475 & 0.488 & 0.413 & 0.351 & 0.312 & 0.225 & 0.164 \\
 & +DDN & 0.713 & 0.449 & 0.375 & 0.360 & 0.339 & 0.241 & 0.158 \\
 & IMP(\%) & 11.20 & 3.48 & 9.50 & 0.13 & 9.03 & 1.03 & 0.94 \\
 \midrule
\multirow{4}{*}{Crossformer} & +WDAN & \textbf{0.407} & \textbf{0.408} & \textbf{0.336} & \textbf{0.349} & \textbf{0.261} & \textbf{0.224} & \textbf{0.159} \\
 & +SAN & 0.417 & 0.450 & 0.370 & 0.373 & 0.308 & 0.225 & 0.262 \\
 & +DDN & 0.431 & 0.434 & 0.350 & 0.358 & 0.270 & 0.242 & 0.165 \\
 & IMP(\%) & 2.37 & 5.91 & 4.00 & 2.48 & 3.34 & 0.56 & 3.69 \\
 \midrule
\multirow{4}{*}{FEDformer} & +WDAN & \textbf{0.441} & \textbf{0.427} & \textbf{0.374} & \textbf{0.354} & \textbf{0.270} & \textbf{0.227} & \textbf{0.170} \\
 & +SAN & 0.449 & 0.493 & 0.388 & 0.360 & 0.277 & 0.232 & 0.197 \\
 & +DDN & 0.499 & 0.472 & 0.433 & 0.374 & 0.271 & 0.243 & 0.170 \\
 & IMP(\%) & 1.92 & 9.64 & 3.52 & 1.89 & 0.23 & 2.09 & 0.38 \\
 \bottomrule
\end{tabular}
\caption{Comparison of multivariate time series forecasting performance between WDAN and mainstream normalization methods across different backbone models. We report the average values of the MSE metric across different prediction horizons. The best results for each backbone model are highlighted in bold. IMP indicates the percentage improvement in MSE of WDAN.}
\label{tab:normalization_simple}
\end{table*}

\begin{table*}[!t]
    \centering
    \begin{tabular}{c|cccccc|cccccc}
    \toprule
    \multirow{2}{*}{Methods} & \multicolumn{6}{c|}{iTransformer} & \multicolumn{6}{c}{Crossformer} \\
    & \multicolumn{2}{c}{WDAN} & \multicolumn{2}{c}{MovingAvg} & \multicolumn{2}{c|}{No Differencing} & \multicolumn{2}{c}{WDAN} & \multicolumn{2}{c}{MovingAvg} & \multicolumn{2}{c}{No Differencing} \\
    \cmidrule(r){2-3}  \cmidrule(r){4-5} \cmidrule(r){6-7} \cmidrule(r){8-9} \cmidrule(r){10-11} \cmidrule(r){12-13}
    Metric & MSE & MAE & MSE & MAE & MSE & MAE & MSE & MAE & MSE & MAE & MSE & MAE \\
    \midrule
    Exchange & \textbf{0.407} & \textbf{0.427} & 0.408 & 0.427 & 0.438 & 0.441 & \textbf{0.407} & \textbf{0.424} & 0.418 & 0.432 & 0.437 & 0.441 \\
    ETTh2 & \textbf{0.334} & \textbf{0.386} & 0.360 & 0.405 & 0.338 & 0.389 & \textbf{0.336} & \textbf{0.387} & 0.371 & 0.410 & 0.342 & 0.390 \\
    Weather & \textbf{0.224} & \textbf{0.267} & 0.231 & 0.272 & 0.227 & 0.270 & \textbf{0.224} & \textbf{0.267} & 0.226 & 0.269 & 0.228 & 0.270 \\
    \bottomrule
    \end{tabular}
    \caption{Results of ablation study for WDAN. We report the average values of the metrcis across different prediction horizons.}
    \label{tab:ablation}
\end{table*}

\subsection{Ablation Study}

To further validate the effectiveness of WDAN, we conduct an ablation study focusing on its two key components: (1) the wavelet-based decomposition and (2) the use of first-order differencing for modeling trend dynamics. We compare our full WDAN framework with two variants: (1) \textbf{MovingAvg}, which replaces the DWT with a simple moving average method \cite{zeng2023transformers} while retaining the differenced trend features; (2) \textbf{NoDiff}, which maintains the DWT-based decomposition but removes the first-order differencing feature from the trend statistics prediction module. We perform experiments using two representative backbone models, iTransformer and Crossformer, across three datasets with varying levels of non-stationarity: Exchange (high), ETTh2 (moderate), and Weather (low).

The results in Table \ref{tab:ablation} demonstrate that WDAN generally outperforms its variants across different settings. On the highly non-stationary Exchange dataset, WDAN consistently outperforms the NoDiff variant across all horizons for both backbones, highlighting the importance of differencing in modeling non-stationary trend dynamics. 
% The performance gap becomes more pronounced for longer horizons ($H=720$), where removing differencing increases MSE and MAE by approximately 10\% with both backbone models. 
On the relatively stationary Weather dataset, all three variants perform similarly, suggesting that WDAN remains effective without introducing unnecessary complexity in more stationary settings.

\section{Conclusion}

In this paper, we present WDAN, a novel normalization framework designed to tackle non-stationarity in time series forecasting. By decomposing input sequences into low-frequency and high-frequency components using wavelet transforms, WDAN explicitly disentangles the sources of non-stationarity and applies component-wise normalization strategies. The proposed framework further enhances adaptability by leveraging differencing techniques and multi-scale residual dynamics to predict future normalization statistics. WDAN is model-agnostic and can be seamlessly integrated with a variety of backbone forecasting models. Extensive experiments across multiple datasets and forecasting models demonstrate the effectiveness and generalizability of WDAN. The success of WDAN underscores the importance of recognizing and separately modeling the diverse non-stationary behaviors exhibited by different time series components. This work opens promising directions for future research, including exploring adaptive wavelet basis selection, extending the approach to capture cross-variable dependencies in multivariate settings, and investigating theoretical connections between wavelet-based decomposition and formal stationarity properties.

\appendix
\section{Experiment Setting}

\subsection{Setting Details}
\label{sec:appendix_implement}

All experiments are conducted on an NVIDIA A100 GPU using PyTorch. We utilize ADAM \cite{kingma2014adam} as the optimizer and employing Mean Squared Error (MSE) as the loss function. We choose coiflet 3 as the default wavelet basis function. In the statistics prediction module, $\text{MLP}_i, i=1,2,3,4$ use single-layer perceptrons for feature mapping by default, while $\text{MLP}_\mu$ and $\text{MLP}_\sigma$ can use multi-layer perceptrons with the number of layers treated as a hyperparameter. For backbone networks, we use the default hyperparameter settings and perform grid search to optimize the additional hyperparameters introduced by WDAN.

\subsection{Backbones}
\label{sec:app_backbones}

Our backbone models are described as follows:

\begin{itemize}
    \item iTransformer \cite{liu2023itransformer} inverts the traditional Transformer architecture by embedding the whole time series of each variate independently into a token and treating each variate as a token for attention mechanism. The source code is available at \url{https://github.com/thuml/iTransformer}.
    \item PatchTST \cite{Yuqietal-2023-PatchTST} is a Transformer-based model that segments time series into subseries-level patches as input tokens rather than using point-wise inputs. It employs channel-independence where each univariate series is processed separately while sharing the same parameters across all series. The source code is available at \url{https://github.com/yuqinie98/PatchTST}.
    \item Crossformer \cite{zhang2022crossformer} introduces a two-stage attention mechanism that separately handles temporal and variate dimensions. It employs a unique cross-dimension attention design with a two-stage embedding architecture that decomposes the complex spatiotemporal modeling task, allowing it to efficiently capture both temporal patterns and inter-variable relationships in multivariate time series. The source code is available at \url{https://github.com/Thinklab-SJTU/Crossformer}.
    \item FEDformer \cite{zhou2022fedformer} combines seasonal-trend decomposition with frequency domain analysis for improved time series forecasting. It uses a mixture of experts to capture global properties while its frequency enhanced blocks extract patterns through Fourier or Wavelet transforms. The source code is available at \url{https://github.com/MAZiqing/FEDformer}.
\end{itemize}

\section{Supplement Experiments Results}

\subsection{Full Multivariate Forecasting Results}
\label{sec:appendix_full_rets}

Due to space limitations in the main text, we provide complete multivariate time series prediction results in this subsection. Table \ref{tab:full_rets_1} and \ref{tab:full_rets_2} show the prediction results of different backbone models under various normalization method frameworks. These results demonstrate that our proposed WDAN achieves leading performance across different datasets and backbone models.

\begin{table*}[htbp]
\centering
{\setlength{\tabcolsep}{1mm}
\begin{tabular}{cc|cccccccc|cccccccc}
\toprule
\multicolumn{2}{c|}{Methods} & \multicolumn{2}{c}{iTransformer} & \multicolumn{2}{c}{+SAN} & \multicolumn{2}{c}{+DDN} & \multicolumn{2}{c|}{\textbf{+WDAN}} & \multicolumn{2}{c}{PatchTST} & \multicolumn{2}{c}{+SAN} & \multicolumn{2}{c}{+DDN} & \multicolumn{2}{c}{\textbf{+WDAN}} \\
 \cmidrule(r){3-4}  \cmidrule(r){5-6} \cmidrule(r){7-8} \cmidrule(r){9-10} \cmidrule(r){11-12} \cmidrule(r){13-14} \cmidrule(r){15-16} \cmidrule(r){17-18}
\multicolumn{2}{c|}{Metric} & MSE & MAE & MSE & MAE & MSE & MAE & MSE & MAE & MSE & MAE & MSE & MAE & MSE & MAE & MSE & MAE \\
\midrule
\multirow{4}{*}{\rotatebox[origin=c]{90}{Exchange}} & 96 & 0.118 & 0.250 & 0.091 & 0.215 & 0.092 & 0.215 & \textbf{0.083} & \textbf{0.203} & 0.101 & 0.230 & 0.093 & 0.217 & 0.087 & 0.207 & \textbf{0.084} & \textbf{0.204} \\
 & 192 & 0.246 & 0.370 & 0.189 & 0.311 & 0.190 & 0.311 & \textbf{0.176} & \textbf{0.300} & 0.200 & 0.326 & 0.192 & 0.314 & 0.177 & \textbf{0.300} & \textbf{0.176} & 0.302 \\
 & 336 & 0.429 & 0.493 & \textbf{0.353} & \textbf{0.429} & 0.365 & 0.437 & 0.361 & 0.436 & 0.405 & 0.470 & 0.378 & 0.447 & 0.449 & 0.487 & \textbf{0.347} & \textbf{0.429} \\
 & 720 & 1.101 & 0.796 & 1.032 & \textbf{0.766} & 1.068 & 0.777 & \textbf{1.009} & 0.767 & 1.463 & 0.904 & 1.238 & 0.826 & 2.139 & 1.114 & \textbf{1.081} & \textbf{0.781} \\
 \midrule
\multirow{4}{*}{\rotatebox[origin=c]{90}{ETTh1}} & 96 & 0.390 & 0.422 & 0.388 & 0.418 & 0.379 & 0.407 & \textbf{0.368} & \textbf{0.397} & 0.399 & 0.419 & 0.399 & 0.416 & 0.399 & 0.417 & \textbf{0.369} & \textbf{0.398} \\
 & 192 & 0.427 & 0.448 & 0.422 & 0.441 & 0.417 & 0.432 & \textbf{0.406} & \textbf{0.421} & 0.447 & 0.450 & 0.449 & 0.447 & 0.419 & 0.429 & \textbf{0.407} & \textbf{0.422} \\
 & 336 & 0.459 & 0.471 & 0.447 & 0.459 & 0.458 & 0.459 & \textbf{0.428} & \textbf{0.444} & 0.472 & 0.474 & 0.495 & 0.481 & 0.459 & 0.455 & \textbf{0.443} & \textbf{0.449} \\
 & 720 & 0.562 & 0.545 & 0.477 & 0.492 & 0.574 & 0.539 & \textbf{0.452} & \textbf{0.470} & 0.519 & 0.511 & 0.611 & 0.545 & 0.521 & \textbf{0.503} & \textbf{0.515} & 0.504 \\
 \midrule
\multirow{4}{*}{\rotatebox[origin=c]{90}{ETTh2}} & 96 & 0.301 & 0.360 & 0.288 & 0.350 & 0.277 & 0.341 & \textbf{0.269} & \textbf{0.334} & 0.322 & 0.376 & 0.341 & 0.386 & 0.281 & 0.341 & \textbf{0.271} & \textbf{0.335} \\
 & 192 & 0.379 & 0.408 & 0.352 & 0.392 & 0.338 & 0.381 & \textbf{0.331} & \textbf{0.376} & 0.412 & 0.432 & 0.437 & 0.447 & 0.395 & 0.407 & \textbf{0.336} & \textbf{0.376} \\
 & 336 & 0.415 & 0.436 & 0.379 & 0.417 & 0.365 & 0.407 & \textbf{0.358} & \textbf{0.403} & 0.450 & 0.455 & 0.434 & 0.456 & 0.374 & 0.414 & \textbf{0.357} & \textbf{0.401} \\
 & 720 & 0.435 & 0.463 & 0.412 & 0.450 & 0.409 & 0.448 & \textbf{0.377} & \textbf{0.429} & 0.497 & 0.486 & 0.440 & 0.461 & 0.451 & 0.479 & \textbf{0.394} & \textbf{0.440} \\
 \midrule
\multirow{4}{*}{\rotatebox[origin=c]{90}{ETTm1}} & 96 & 0.315 & 0.368 & 0.296 & \textbf{0.348} & 0.308 & 0.352 & \textbf{0.291} & 0.350 & 0.296 & 0.354 & 0.297 & \textbf{0.347} & 0.299 & 0.351 & \textbf{0.293} & 0.348 \\
 & 192 & 0.345 & 0.386 & 0.330 & \textbf{0.369} & 0.340 & 0.371 & \textbf{0.330} & 0.375 & 0.346 & 0.388 & \textbf{0.331} & \textbf{0.368} & 0.336 & 0.374 & 0.335 & 0.376 \\
 & 336 & 0.379 & 0.407 & \textbf{0.362} & \textbf{0.388} & 0.370 & 0.389 & 0.362 & 0.396 & 0.400 & 0.421 & 0.363 & \textbf{0.388} & 0.366 & 0.394 & \textbf{0.362} & 0.397 \\
 & 720 & 0.443 & 0.444 & \textbf{0.412} & 0.418 & 0.422 & \textbf{0.417} & 0.412 & 0.422 & 0.477 & 0.464 & 0.415 & 0.418 & 0.437 & 0.439 & \textbf{0.413} & \textbf{0.411} \\
 \midrule
\multirow{4}{*}{\rotatebox[origin=c]{90}{ETTm2}} & 96 & 0.182 & 0.276 & 0.176 & 0.268 & 0.163 & 0.254 & \textbf{0.162} & \textbf{0.253} & 0.179 & 0.272 & 0.184 & 0.274 & 0.174 & 0.258 & \textbf{0.172} & \textbf{0.257} \\
 & 192 & 0.244 & 0.317 & 0.234 & 0.307 & \textbf{0.217} & \textbf{0.291} & 0.217 & 0.292 & 0.248 & 0.317 & 0.263 & 0.324 & 0.262 & 0.317 & \textbf{0.240} & \textbf{0.304} \\
 & 336 & 0.298 & 0.351 & 0.295 & 0.347 & \textbf{0.269} & \textbf{0.327} & 0.273 & 0.331 & 0.311 & 0.357 & 0.317 & 0.361 & 0.332 & 0.351 & \textbf{0.282} & \textbf{0.334} \\
 & 720 & 0.376 & 0.401 & 0.368 & 0.396 & 0.355 & 0.385 & \textbf{0.353} & \textbf{0.382} & \textbf{0.418} & \textbf{0.426} & 0.485 & 0.453 & 0.587 & 0.482 & 0.442 & 0.432 \\
 \midrule
\multirow{4}{*}{\rotatebox[origin=c]{90}{Weather}} & 96 & 0.177 & 0.229 & 0.151 & 0.207 & 0.153 & 0.214 & \textbf{0.147} & \textbf{0.199} & 0.153 & 0.208 & 0.148 & 0.204 & 0.158 & 0.211 & \textbf{0.148} & \textbf{0.202} \\
 & 192 & 0.226 & 0.268 & \textbf{0.195} & 0.249 & 0.197 & 0.258 & 0.196 & \textbf{0.248} & 0.206 & 0.261 & 0.194 & 0.251 & 0.222 & 0.266 & \textbf{0.194} & \textbf{0.248} \\
 & 336 & 0.285 & 0.309 & 0.246 & 0.292 & 0.250 & 0.303 & \textbf{0.242} & \textbf{0.282} & 0.243 & 0.286 & 0.245 & \textbf{0.286} & 0.253 & 0.303 & \textbf{0.242} & 0.290 \\
 & 720 & 0.359 & 0.361 & \textbf{0.311} & 0.341 & 0.326 & 0.370 & 0.312 & \textbf{0.337} & 0.315 & 0.336 & 0.312 & 0.335 & 0.330 & 0.354 & \textbf{0.306} & \textbf{0.332} \\
 \midrule
\multirow{4}{*}{\rotatebox[origin=c]{90}{Electricity}} & 96 & 0.135 & 0.232 & 0.131 & 0.230 & 0.129 & 0.226 & \textbf{0.128} & \textbf{0.225} & 0.137 & 0.241 & 0.137 & 0.242 & 0.129 & 0.226 & \textbf{0.128} & \textbf{0.224} \\
 & 192 & 0.157 & 0.253 & 0.147 & 0.246 & 0.146 & 0.244 & \textbf{0.146} & \textbf{0.242} & 0.153 & 0.256 & 0.152 & 0.256 & 0.146 & 0.242 & \textbf{0.145} & \textbf{0.240} \\
 & 336 & 0.171 & 0.269 & 0.157 & 0.258 & \textbf{0.155} & \textbf{0.255} & 0.156 & 0.256 & 0.169 & 0.272 & 0.167 & 0.271 & 0.161 & 0.260 & \textbf{0.160} & \textbf{0.258} \\
 & 720 & 0.196 & 0.290 & 0.185 & 0.285 & \textbf{0.182} & 0.281 & 0.182 & \textbf{0.281} & 0.205 & 0.302 & 0.200 & 0.300 & 0.196 & 0.293 & \textbf{0.194} & \textbf{0.291} \\
 \bottomrule
\end{tabular}
}
\caption{Full multivariate time series prediction results with iTransformer and PatchTST as backbones under different normalization frameworks. The best results for each backbone model are highlighted in bold.}
\label{tab:full_rets_1}
\end{table*}

\begin{table*}[htbp]
\centering
{\setlength{\tabcolsep}{1mm}
\begin{tabular}{cc|cccccccc|cccccccc}
\toprule
 \multicolumn{2}{c|}{Methods} & \multicolumn{2}{c}{Crossformer} & \multicolumn{2}{c}{+SAN} & \multicolumn{2}{c}{+DDN} & \multicolumn{2}{c|}{\textbf{+WDAN}} & \multicolumn{2}{c}{FEDformer} & \multicolumn{2}{c}{+SAN} & \multicolumn{2}{c}{+DDN} & \multicolumn{2}{c}{\textbf{+WDAN}} \\
 \cmidrule(r){3-4}  \cmidrule(r){5-6} \cmidrule(r){7-8} \cmidrule(r){9-10} \cmidrule(r){11-12} \cmidrule(r){13-14} \cmidrule(r){15-16} \cmidrule(r){17-18}
\multicolumn{2}{c|}{Metric} & MSE & MAE & MSE & MAE & MSE & MAE & MSE & MAE & MSE & MAE & MSE & MAE & MSE & MAE & MSE & MAE \\
\midrule
\multirow{4}{*}{\rotatebox[origin=c]{90}{Exchange}} & 96 & 0.561 & 0.547 & 0.092 & 0.216 & 0.098 & 0.223 & \textbf{0.084} & \textbf{0.205} & 0.744 & 0.668 & 0.102 & 0.230 & 0.107 & 0.235 & \textbf{0.087} & \textbf{0.205} \\
 & 192 & 0.798 & 0.683 & 0.188 & 0.310 & 0.197 & 0.317 & \textbf{0.174} & \textbf{0.300} & 0.928 & 0.760 & 0.226 & 0.342 & 0.280 & 0.377 & \textbf{0.180} & \textbf{0.305} \\
 & 336 & 1.455 & 0.975 & \textbf{0.342} & \textbf{0.422} & 0.389 & 0.453 & 0.350 & 0.431 & 1.180 & 0.857 & \textbf{0.370} & \textbf{0.447} & 0.413 & 0.468 & 0.386 & 0.456 \\
 & 720 & 2.050 & 1.163 & 1.045 & 0.773 & 1.038 & 0.773 & \textbf{1.019} & \textbf{0.762} & 1.901 & 1.068 & \textbf{1.099} & 0.795 & 1.197 & 0.819 & 1.110 & \textbf{0.789} \\
 \midrule
\multirow{4}{*}{\rotatebox[origin=c]{90}{ETTh1}} & 96 & 0.391 & 0.431 & 0.386 & 0.407 & 0.377 & 0.403 & \textbf{0.366} & \textbf{0.396} & 0.486 & 0.498 & 0.448 & 0.458 & 0.392 & 0.418 & \textbf{0.381} & \textbf{0.415} \\
 & 192 & 0.490 & 0.493 & 0.431 & 0.434 & 0.413 & 0.423 & \textbf{0.401} & \textbf{0.415} & 0.522 & 0.512 & 0.484 & 0.477 & 0.440 & 0.447 & \textbf{0.417} & \textbf{0.438} \\
 & 336 & 0.939 & 0.758 & 0.483 & 0.470 & 0.461 & 0.456 & \textbf{0.420} & \textbf{0.429} & 0.545 & 0.534 & 0.502 & 0.490 & 0.479 & 0.469 & \textbf{0.436} & \textbf{0.452} \\
 & 720 & 1.080 & 0.796 & 0.501 & 0.494 & 0.484 & 0.483 & \textbf{0.446} & \textbf{0.468} & 0.678 & 0.620 & 0.539 & 0.517 & 0.578 & 0.519 & \textbf{0.473} & \textbf{0.484} \\
 \midrule
\multirow{4}{*}{\rotatebox[origin=c]{90}{ETTh2}} & 96 & 1.225 & 0.764 & 0.284 & 0.348 & 0.274 & 0.338 & \textbf{0.269} & \textbf{0.336} & 0.405 & 0.459 & 0.301 & 0.361 & 0.306 & 0.369 & \textbf{0.277} & \textbf{0.342} \\
 & 192 & 1.253 & 0.810 & 0.343 & 0.384 & 0.339 & 0.380 & \textbf{0.333} & \textbf{0.378} & 0.431 & 0.480 & 0.370 & 0.407 & 0.385 & 0.419 & \textbf{0.347} & \textbf{0.386} \\
 & 336 & 1.500 & 0.920 & 0.378 & 0.413 & 0.365 & 0.407 & \textbf{0.359} & \textbf{0.405} & 0.436 & 0.481 & 0.405 & 0.435 & 0.443 & 0.456 & \textbf{0.391} & \textbf{0.423} \\
 & 720 & 3.825 & 1.606 & 0.476 & 0.481 & 0.424 & 0.455 & \textbf{0.384} & \textbf{0.428} & 0.502 & 0.514 & \textbf{0.476} & \textbf{0.477} & 0.599 & 0.527 & 0.483 & 0.478 \\
 \midrule
\multirow{4}{*}{\rotatebox[origin=c]{90}{ETTm1}} & 96 & 0.348 & 0.389 & 0.295 & \textbf{0.346} & \textbf{0.292} & 0.349 & 0.293 & 0.349 & 0.435 & 0.457 & 0.306 & 0.358 & 0.314 & 0.370 & \textbf{0.295} & \textbf{0.354} \\
 & 192 & 0.360 & 0.395 & \textbf{0.332} & \textbf{0.368} & 0.336 & 0.372 & 0.332 & 0.373 & 0.432 & 0.455 & 0.338 & \textbf{0.378} & 0.342 & 0.386 & \textbf{0.330} & 0.378 \\
 & 336 & 0.586 & 0.573 & 0.362 & \textbf{0.387} & 0.366 & 0.392 & \textbf{0.360} & 0.394 & 0.454 & 0.470 & 0.373 & \textbf{0.398} & 0.380 & 0.405 & \textbf{0.370} & 0.403 \\
 & 720 & 0.977 & 0.787 & 0.504 & 0.459 & 0.439 & 0.431 & \textbf{0.413} & \textbf{0.411} & 0.499 & 0.489 & 0.426 & 0.429 & 0.461 & 0.443 & \textbf{0.420} & \textbf{0.425} \\
 \midrule
\multirow{4}{*}{\rotatebox[origin=c]{90}{ETTm2}} & 96 & 0.310 & 0.388 & 0.181 & 0.273 & 0.163 & 0.254 & \textbf{0.162} & \textbf{0.254} & 0.317 & 0.379 & 0.178 & 0.271 & 0.172 & 0.262 & \textbf{0.167} & \textbf{0.258} \\
 & 192 & 0.387 & 0.438 & 0.286 & 0.343 & 0.224 & 0.298 & \textbf{0.216} & \textbf{0.290} & 0.338 & 0.390 & 0.239 & 0.314 & \textbf{0.234} & 0.305 & 0.237 & \textbf{0.299} \\
 & 336 & 0.614 & 0.558 & 0.334 & 0.376 & 0.287 & 0.343 & \textbf{0.278} & \textbf{0.336} & 0.366 & 0.404 & 0.306 & 0.363 & 0.288 & 0.345 & \textbf{0.282} & \textbf{0.338} \\
 & 720 & 1.279 & 0.813 & 0.429 & 0.436 & 0.409 & 0.431 & \textbf{0.390} & \textbf{0.405} & 0.421 & 0.446 & \textbf{0.386} & 0.418 & 0.389 & \textbf{0.411} & 0.395 & 0.417 \\
 \midrule
\multirow{4}{*}{\rotatebox[origin=c]{90}{Weather}} & 96 & 0.173 & 0.244 & 0.150 & 0.207 & 0.153 & 0.210 & \textbf{0.147} & \textbf{0.198} & 0.317 & 0.373 & 0.154 & 0.212 & 0.154 & 0.213 & \textbf{0.153} & \textbf{0.207} \\
 & 192 & 0.218 & 0.288 & 0.195 & 0.250 & 0.200 & 0.260 & \textbf{0.195} & \textbf{0.248} & 0.347 & 0.397 & 0.200 & 0.256 & 0.201 & 0.260 & \textbf{0.193} & \textbf{0.248} \\
 & 336 & 0.266 & 0.328 & 0.244 & 0.291 & 0.260 & 0.308 & \textbf{0.244} & \textbf{0.287} & 0.354 & 0.392 & 0.252 & 0.296 & 0.256 & 0.305 & \textbf{0.247} & \textbf{0.293} \\
 & 720 & 0.334 & 0.380 & 0.311 & 0.341 & 0.355 & 0.370 & \textbf{0.310} & \textbf{0.336} & 0.400 & 0.421 & 0.321 & 0.348 & 0.362 & 0.370 & \textbf{0.313} & \textbf{0.341} \\
 \midrule
\multirow{4}{*}{\rotatebox[origin=c]{90}{Electricity}} & 96 & 0.137 & 0.238 & 0.138 & 0.237 & 0.137 & 0.237 & \textbf{0.128} & \textbf{0.223} & 0.227 & 0.339 & 0.170 & 0.278 & 0.147 & 0.250 & \textbf{0.142} & \textbf{0.245} \\
 & 192 & 0.157 & 0.258 & 0.161 & 0.258 & 0.155 & 0.254 & \textbf{0.148} & \textbf{0.244} & 0.231 & 0.343 & 0.186 & 0.293 & 0.161 & \textbf{0.263} & \textbf{0.159} & 0.264 \\
 & 336 & 0.193 & 0.294 & 0.177 & 0.275 & 0.169 & 0.270 & \textbf{0.164} & \textbf{0.262} & 0.259 & 0.366 & 0.202 & 0.309 & \textbf{0.170} & \textbf{0.273} & 0.174 & 0.279 \\
 & 720 & 0.281 & 0.358 & 0.572 & 0.576 & 0.199 & 0.300 & \textbf{0.196} & \textbf{0.294} & 0.297 & 0.393 & 0.232 & 0.334 & \textbf{0.203} & \textbf{0.303} & 0.203 & 0.309 \\
 \bottomrule
\end{tabular}
}
\caption{Full multivariate time series prediction results with Crossformer and FEDformer as backbones under different normalization frameworks. The best results for each backbone model are highlighted in bold.}
\label{tab:full_rets_2}
\end{table*}

\subsection{Analysis on Various Input Lengths}

\begin{figure*}[htbp]
    \centering
    \includegraphics[width=1\linewidth]{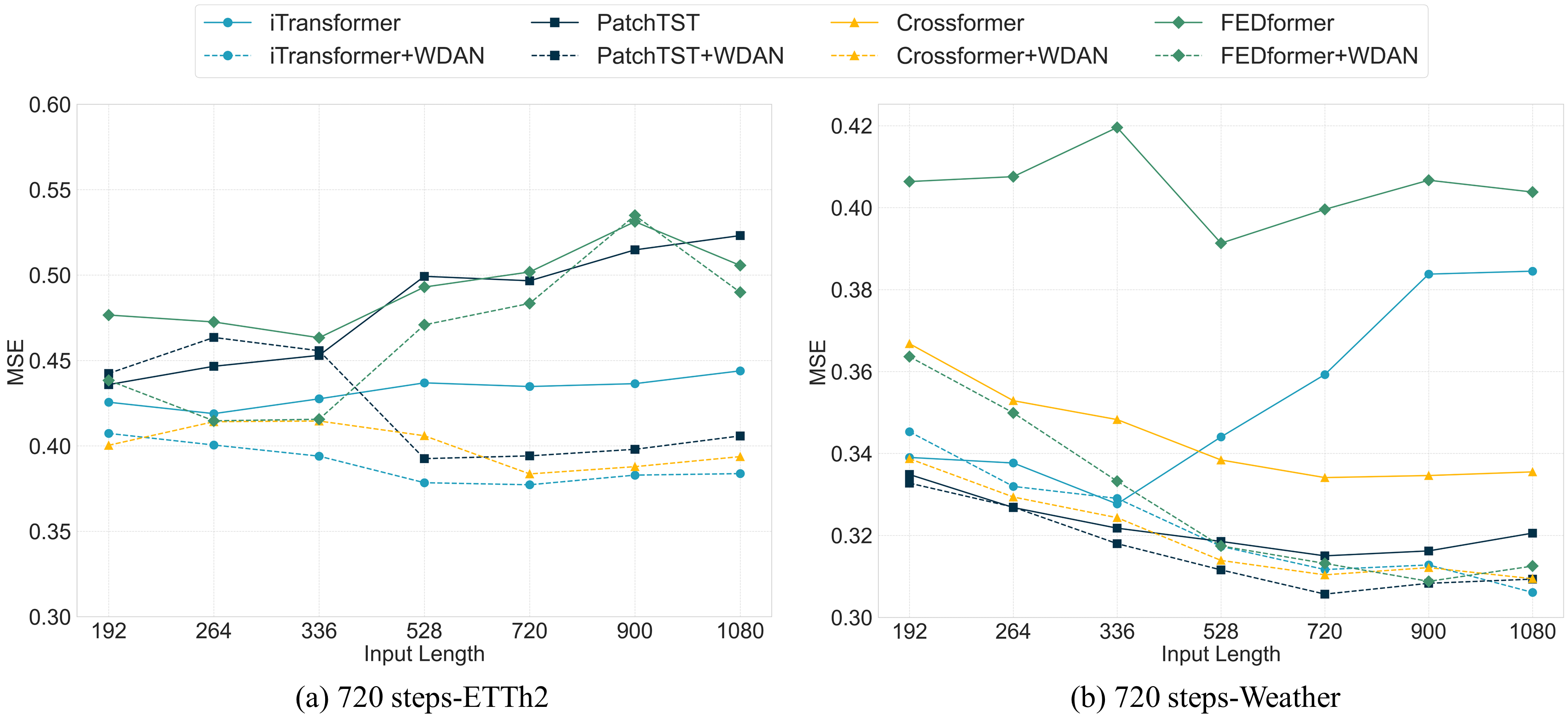}
    \caption{The long-term forecasting MSE evaluations of different backbone models under various input lengths. Large values are discarded to illustrate the overall trend better.}
    \label{fig:sen_seq_len}
\end{figure*}

The length of input sequences is a crucial hyperparameter in time series forecasting tasks as it determines the amount of historical information available for deep models to mine. Ideally, a well-designed model should benefit from longer input sequences, leading to improved forecasting performance. However, recent studies have shown that many Transformer-based deep models struggle with long input sequences \cite{zeng2023transformers}. In fact, their forecasting accuracy tends to degrade as the input length increases. This phenomenon has been attributed to the non-stationarity of time series data \cite{liu2023adaptive}. As the input length grows, the distributional discrepancy across time points become more pronounced, making it harder for deep models to learn consistent temporal patterns. Therefore, as a model-agnostic framework for addressing non-stationarity, we expect that after processing non-stationarity with WDAN, the prediction errors of deep models would decrease or at least remain stable with increasing sequence length.

To validate our hypothesis, we evaluated the long-term ($H=720$) forecasting performance of various backbone models under different input lengths $L\in\{192, 264, 336, 528, 720, 900, 1080\}$. The MSE evaluation results are presented in Fig. \ref{fig:sen_seq_len}. We can observe that with the assistance of WDAN, the predictive performance of deep backbone models is significantly enhanced across different input lengths. Without WDAN, most models suffer from performance degradation as input length increases across both datasets. In contrast, with WDAN, this degradation is substantially mitigated, and the performance of most backbone models improves as the input length grows. To be specific, on the Weather dataset, iTransformer achieves a reduction on MSE of 11.36\% when prolonging input from 192 steps to 1080 steps, and the average improvement on four backbones is 10.28\%. These results greatly meet our expectations and also validate the effectiveness of WDAN on various input lengths.

\subsection{Hyperparameter Sensitivity}

\begin{figure*}[htbp]
    \centering
    \includegraphics[width=1\linewidth]{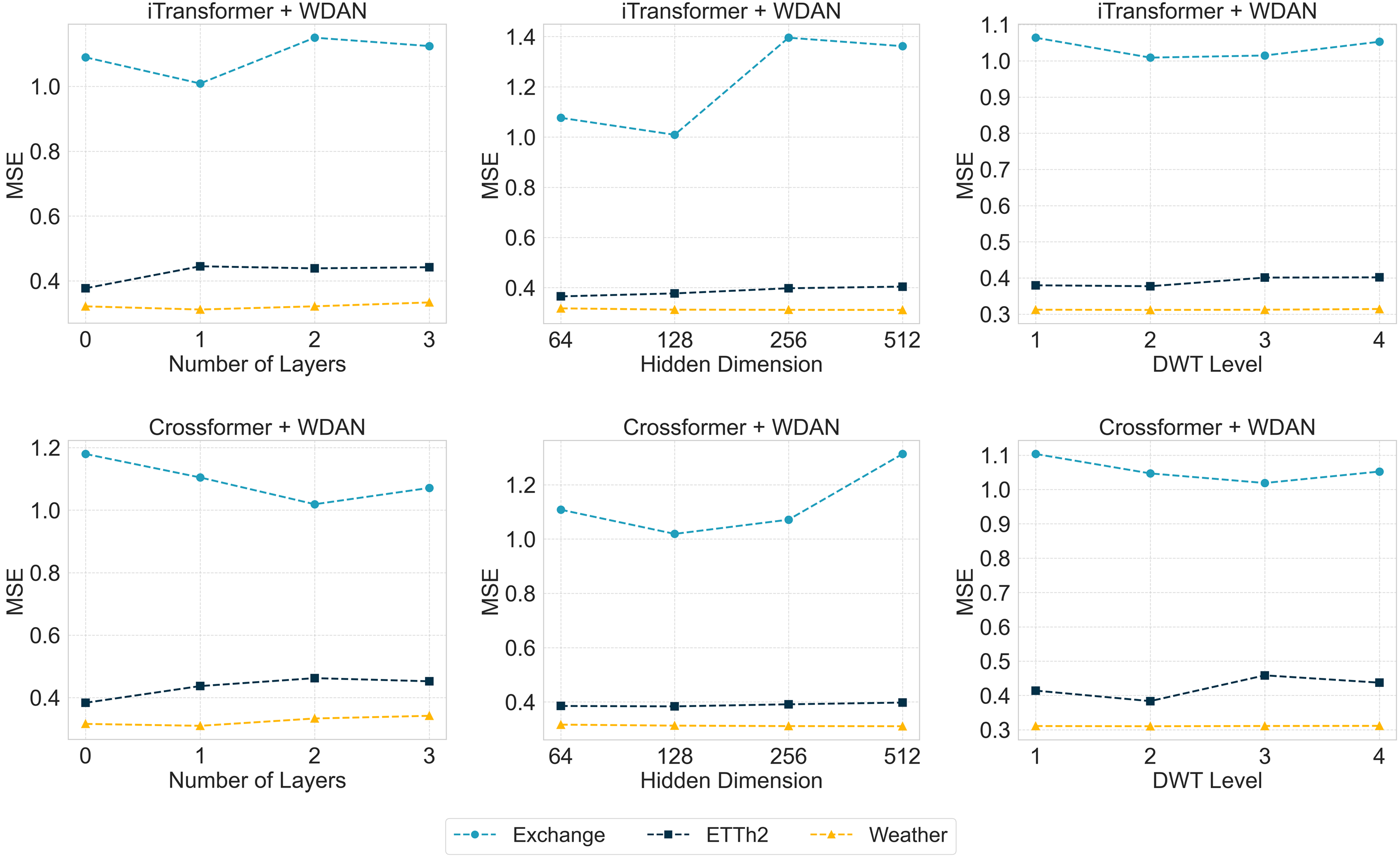}
    \caption{Hyperparameter sensitivity with respect to the number of MLP layers of statistics prediction module, the hidden dimension and the level of DWT. The results are recorded with the lookback window length $T=720$ and the forecast window length $H=720$.}
    \label{fig:sensitivity}
\end{figure*}

We investigate the sensitivity of WDAN to three key hyperparameters: the number of MLP layers in the statistics prediction module, the hidden dimension and the level of the DWT. The results of the sensitivity analysis are shown in Fig. \ref{fig:sensitivity}. We observe that in the highly non-stationary Exchange dataset, different hyperparameter settings significantly impact prediction results. It suggests that careful hyperparameter tuning is particularly important for non-stationary datasets. In contrast, on other relatively stationary datasets, the model performance remains relatively robust across different hyperparameter settings.

\subsection{Training Strategy Ablation}

In existing normalization methods, different training strategies have been adopted for the statistics prediction module and the backbone model. To validate the effectiveness of our three-stage training strategy, we conducted ablation experiments comparing it with other training approaches found in existing work. Specifically, we compared four training strategies: (1) Three-stage, the approach adopted in this paper and in DDN \cite{dai2024ddn}; (2) Two-stage (Alternative): alternately training the statistical prediction module and the backbone model, freezing the parameters of the statistical prediction module when training the backbone; (3) Two-stage (Co-train): training the statistical prediction module in the first stage, then simultaneously training both modules in the second stage; (4) Single-stage: directly training both modules simultaneously.

Table \ref{tab:ablation_strategy} show the results of these ablation experiments, showing that the three-stage strategy outperforms other training strategies in most settings. These results indicate that the three-stage training strategy offers greater training flexibility while maintaining a strong lower bound on predictive performance.

\begin{table*}[htbp]
\centering
\begin{tabular}{cc|cccccccc}
\toprule
\multicolumn{2}{c|}{\multirow{2}{*}{Methods}} & \multicolumn{8}{c}{iTransformer + WDAN} \\
\multicolumn{2}{c|}{} & \multicolumn{2}{c}{Three-stage} & \multicolumn{2}{c}{Two-stage (Alternative)} & \multicolumn{2}{c}{Two-stage (Co-train)} & \multicolumn{2}{c}{Single-stage} \\
 \cmidrule(r){3-4}  \cmidrule(r){5-6} \cmidrule(r){7-8} \cmidrule(r){9-10}
\multicolumn{2}{c|}{Metric} & MSE & MAE & MSE & MAE & MSE & MAE & MSE & MAE \\
\midrule
\multirow{4}{*}{\rotatebox[origin=c]{90}{Exchange}} & 96 & \textbf{0.083} & \textbf{0.203} & 0.094 & 0.217 & 0.083 & 0.203 & 0.094 & 0.219 \\
 & 192 & \textbf{0.176} & \textbf{0.300} & 0.188 & 0.312 & 0.177 & 0.302 & 0.201 & 0.324 \\
 & 336 & 0.361 & 0.436 & \textbf{0.361} & \textbf{0.436} & 0.369 & 0.437 & 0.424 & 0.475 \\
 & 720 & \textbf{1.009} & \textbf{0.767} & 1.115 & 0.811 & 1.009 & 0.767 & 1.139 & 0.819 \\
 \midrule
\multirow{4}{*}{\rotatebox[origin=c]{90}{ETTh2}} & 96 & \textbf{0.269} & \textbf{0.334} & 0.276 & 0.340 & \textbf{0.269} & \textbf{0.334} & 0.274 & 0.339 \\
 & 192 & \textbf{0.331} & \textbf{0.376} & 0.342 & 0.382 & \textbf{0.331} & \textbf{0.376} & 0.336 & 0.381 \\
 & 336 & \textbf{0.358} & \textbf{0.403} & 0.359 & 0.405 & 0.365 & 0.407 & 0.370 & 0.411 \\
 & 720 & \textbf{0.377} & \textbf{0.429} & 0.385 & 0.436 & 0.401 & 0.442 & 0.396 & 0.440 \\
 \midrule
\multirow{4}{*}{\rotatebox[origin=c]{90}{Weather}} & 96 & \textbf{0.147} & \textbf{0.199} & 0.148 & 0.204 & 0.150 & 0.202 & 0.150 & 0.206 \\
 & 192 & 0.200 & \textbf{0.251} & 0.198 & 0.254 & 0.209 & 0.259 & \textbf{0.197} & 0.257 \\
 & 336 & \textbf{0.242} & \textbf{0.282} & 0.245 & 0.291 & 0.253 & 0.292 & 0.252 & 0.304 \\
 & 720 & \textbf{0.312} & \textbf{0.337} & 0.313 & 0.342 & 0.319 & 0.340 & 0.322 & 0.350 \\
 \midrule
 \multicolumn{2}{c|}{\multirow{2}{*}{Methods}} & \multicolumn{8}{c}{Crossformer + WDAN} \\
\multicolumn{2}{c|}{} & \multicolumn{2}{c}{Three-stage} & \multicolumn{2}{c}{Two-stage (Alternative)} & \multicolumn{2}{c}{Two-stage (Co-train)} & \multicolumn{2}{c}{Single-stage} \\
 \cmidrule(r){3-4}  \cmidrule(r){5-6} \cmidrule(r){7-8} \cmidrule(r){9-10}
\multicolumn{2}{c|}{Metric} & MSE & MAE & MSE & MAE & MSE & MAE & MSE & MAE \\
\midrule
\multirow{4}{*}{\rotatebox[origin=c]{90}{Exchange}} & 96 & \textbf{0.084} & \textbf{0.205} & 0.091 & 0.214 & \textbf{0.084} & \textbf{0.205} & 0.095 & 0.220 \\
 & 192 & \textbf{0.174} & \textbf{0.299} & 0.191 & 0.314 & \textbf{0.174} & \textbf{0.299} & 0.204 & 0.326 \\
 & 336 & \textbf{0.358} & \textbf{0.437} & 0.373 & 0.452 & 0.364 & 0.440 & 0.402 & 0.469 \\
 & 720 & \textbf{1.067} & \textbf{0.788} & 1.176 & 0.838 & 1.172 & 0.817 & 1.123 & 0.816 \\
 \midrule
\multirow{4}{*}{\rotatebox[origin=c]{90}{ETTh2}} & 96 & \textbf{0.269} & \textbf{0.337} & 0.272 & 0.337 & 0.276 & 0.337 & 0.276 & 0.344 \\
 & 192 & \textbf{0.333} & \textbf{0.377} & 0.334 & 0.381 & 0.344 & 0.383 & 0.351 & 0.390 \\
 & 336 & \textbf{0.359} & \textbf{0.405} & 0.366 & 0.411 & 0.371 & 0.408 & 0.379 & 0.418 \\
 & 720 & \textbf{0.379} & \textbf{0.425} & 0.397 & 0.436 & 0.391 & 0.433 & 0.415 & 0.450 \\
 \midrule
\multirow{4}{*}{\rotatebox[origin=c]{90}{Weather}} & 96 & \textbf{0.147} & \textbf{0.198} & 0.149 & 0.205 & 0.152 & 0.202 & 0.150 & 0.205 \\
 & 192 & 0.207 & 0.259 & 0.200 & 0.258 & 0.221 & 0.269 & \textbf{0.197} & \textbf{0.253} \\
 & 336 & \textbf{0.243} & \textbf{0.285} & 0.243 & 0.295 & 0.264 & 0.300 & 0.248 & 0.297 \\
 & 720 & \textbf{0.310} & \textbf{0.336} & 0.313 & 0.341 & 0.321 & 0.341 & 0.322 & 0.352 \\
 \bottomrule
\end{tabular}
\caption{Results of training strategy ablation study with iTransformer and Crossformer as the backbone models. The best results are highlighted in bold.}
\label{tab:ablation_strategy}
\end{table*}

\bibliography{reference}

\end{document}